\pdfoutput=1
\documentclass[10pt, conference, letterpaper, english]{IEEEtran}
\IEEEoverridecommandlockouts

\usepackage{	latexsym,%
		    	amsmath,%
			amssymb,%
			amsthm,
			graphicx,
			relsize,
			caption,
			accents,
			caption,
			subcaption,
			tikz,
			pgfplots,
			bm,
			mathtools}

\usepackage[section]{placeins}
\usepackage[inline]{enumitem}
\usepackage[utf8]{inputenc}
\usepackage{array}
\usepackage{cite}
\usepackage[ruled,lined]{algorithm2e}

\pgfplotsset{compat=1.5.1}
\usetikzlibrary{patterns}
\usetikzlibrary{shapes.geometric}
\usetikzlibrary{shapes,snakes,patterns}
\usetikzlibrary{arrows,positioning,automata,calc}
\usetikzlibrary{plotmarks}

\usepackage{microtype}
\usepackage{graphicx}
\usepackage{booktabs} 

\usepackage{hyperref}

\definecolor{bleudefrance}{rgb}{0.19, 0.55, 0.91}

\newcommand{\R}{\mathbb{R}}

\newcommand{\ip}[2]{\ensuremath{\left\langle #1,#2\right\rangle}}

\newcommand{\ghat}{\hat{\mathbf{g}}}
\newcommand{\bw}{\mathbf{w}}
\newcommand{\by}{\mathbf{y}}

\newcolumntype{L}[1]{>{\raggedright\let\newline\\\arraybackslash\hspace{0pt}}m{#1}}
\newcolumntype{C}[1]{>{\centering\let\newline\\\arraybackslash\hspace{0pt}}m{#1}}
\newcolumntype{R}[1]{>{\raggedleft\let\newline\\\arraybackslash\hspace{0pt}}m{#1}}

\newtheorem{theorem}{Theorem}
\newtheorem{example}{Example}
\allowdisplaybreaks[2]

\newtheorem{lemma}{Lemma}

\newtheorem{proposition}{Proposition}

\begin{document}

\title{Adaptive Distributed Stochastic Gradient Descent for Minimizing Delay in the Presence of Stragglers}

\author{
\IEEEauthorblockN{Serge Kas Hanna\IEEEauthorrefmark{1}, Rawad Bitar\IEEEauthorrefmark{1}, Parimal Parag\IEEEauthorrefmark{2}, Venkat Dasari\IEEEauthorrefmark{3}, and  Salim El Rouayheb\IEEEauthorrefmark{1} }
\IEEEauthorblockA{
\IEEEauthorrefmark{1} Department of Electrical and Computer Engineering, Rutgers University, Piscataway, NJ, USA \\
\IEEEauthorrefmark{2} Department of Electrical Communication Engineering, Indian Institute of Science, Bengaluru, KA, India  \\
\IEEEauthorrefmark{3} US Army Research Laboratory, Aberdeen Proving Ground, MD, USA  \\
Emails: \{serge.k.hanna, rawad.bitar, salim.elrouayheb\}@rutgers.edu, parimal@iisc.ac.in, venkateswara.r.dasari.civ@mail.mil
}
\vspace{-1cm}
\thanks{
The work of the first and last authors was supported in part by ARL Grant W911NF-17-1-0032. 

The work of the third author was supported in part by 
the Science and Engineering Research Board under Grant~DSTO-1677, 
the Department of Telecommunications, Government of India, under Grant DOTC-0001, 
the Robert Bosch Center for Cyber-Physical Systems, 
and the Centre for Networked Intelligence (a Cisco CSR initiative) of the Indian Institute of Science, Bangalore. 
}
}
\maketitle 
\begin{abstract}
We consider the setting where a master wants to run a distributed stochastic gradient descent (SGD) algorithm on $n$ workers each having a subset of the data. Distributed SGD may suffer from the effect of stragglers, i.e., slow or unresponsive workers who cause delays. One solution studied in the literature is to wait at each iteration for the responses of the fastest $k<n$ workers before updating the model, where $k$ is a fixed parameter. The choice of the value of $k$ presents a trade-off between the runtime (i.e., convergence rate) of SGD and the error of the model. Towards optimizing the error-runtime trade-off, we investigate distributed SGD with adaptive~$k$. We first design an adaptive policy for varying $k$ that optimizes this trade-off based on an upper bound on the error as a function of the wall-clock time which we derive. Then, we propose an algorithm for adaptive distributed SGD that is based on a statistical heuristic. We implement our algorithm and provide numerical simulations which confirm our intuition and theoretical analysis.  
\end{abstract}

\section{Introduction}
\label{Intro}
We consider a distributed computation setting in which a master wants to learn a model on a large amount of data in his possession by dividing the computations on $n$ workers. The data at the master consists of a matrix $X\in \R^{m\times d}$ representing $m$ data vectors $\mathbf{x}_\ell$, $\ell = 1,\dots,m$, and a vector $\mathbf{y} \in \R^{m}$ representing the labels of the rows of $X$. Define $A\triangleq [X|\mathbf{y}]$ to be the concatenation of $X$ and $\by$. The master would like to find a model $\bw^\star\in \R^d$ that minimizes a loss function $F(A, \bw)$, i.e,
$
\bw^\star = \arg\min_{\bw \in \R^d} F(A, \bw).$
This optimization problem can be solved using Gradient Descent (GD), which is an iterative algorithm that consists of the following update at each iteration~$j$, 
\begin{equation}\label{eq:update}
\bw_{j+1} = \bw_j -\eta \nabla F(A, \bw_j) \triangleq   \bw_j -  \dfrac{\eta}{m} \sum_{\ell=1}^m \nabla F(\mathbf{a}_\ell, \bw_j) ,
\end{equation}
where $\eta$ is the step size, and $\nabla F(A,\bw)$ is the gradient of $F(A,\bw)$. To distribute the computations, the master partitions the data equally to $n$ workers. The data partitioning is horizontal, i.e., each worker receives a set of rows of $A$ with all their corresponding columns. Let  $S_i$ be the sub-matrix of $A$ sent to worker $i$. Each worker computes a partial gradient defined as
$ 
\nabla F(S_i, \bw_j) \triangleq \dfrac{1}{s}\sum_{\mathbf{a}_\ell \in S_i} \nabla F(\mathbf{a}_\ell, \bw_j),
$
where $s = m/n$ is the number of rows in $S_i$ (assuming $n$ divides $m$). The master computes the average of the received partial gradients to obtain the actual gradient $\nabla F(A,\bw)$, and then updates $\bw_j$. 

In this setting, waiting for the partial gradients of all the workers slows down the process as the master has to wait for the stragglers \cite{DB13}, i.e., slow or unresponsive workers, in order to update $\bw_j$. Many approaches have been proposed in the literature to alleviate the problem of stragglers. A natural approach is to simply ignore the stragglers and obtain an estimate of the gradient rather than the full gradient, see \cite{Gauri,CPMBJ16}. This framework emerges from single-node (non-distributed) mini batch stochastic gradient descent (SGD) \cite{RM51}. Batch SGD is a relaxation of GD in which $\bw_j$ is updated based on a subset (batch) $B$ $(|B|<m)$ of data vectors that is chosen uniformly at random from the set of all $m$ data vectors, i.e.,
$\bw_{j+1} =  \bw_j - \dfrac{\eta}{|B|}\sum_{\mathbf{a}_\ell \in B} F(\mathbf{a}_\ell, \bw_j).$
It is shown that SGD converges to $\bw^\star$ under mild assumptions on the loss function $F(A,\bw)$, but may require a larger number of iterations as compared to GD \cite{Bottou,RM51,BT89,CSSS11,AD11,DGSX12,SS14}. 

Consider the approach where the master updates the model based on the responses of the fastest $k<n$ workers and ignores the remaining stragglers. Henceforth, we call this approach {\em fastest-k SGD}. The update rule for fastest-$k$ SGD is given by
\begin{equation}\label{eeq6}
\mathbf{w}_{j+1}=\mathbf{w}_j-\frac{\eta}{k  }  \sum_{i \in R_j} \nabla F(S_{i},\mathbf{w}_j)\triangleq \mathbf{w}_j-\eta ~\ghat(\mathbf{w}_j) ,
\end{equation} 
where $R_j$ is the set of the fastest $k$ workers at iteration~$j$; and $\ghat(\mathbf{w}_j)$ is the average of the partial gradients received by the master at iteration $j$ which is an estimate of the full gradient $\nabla F(A,\mathbf{w}_j)$. Note that if we assume that the response times of the workers are random {\em iid}, then one can easily show that fastest-$k$ SGD is essentially equivalent to single-node batch SGD since the master updates the model at each iteration based on a uniformly random batch of data vectors belonging to the set of the fastest $k$ workers. Therefore, fastest-$k$ SGD converges to $\bw^\star$ under the random {\em iid} assumption on the response times of the workers and the standard assumptions on the loss function $F(A,\bw)$.

The convergence rate of distributed SGD depends on two factors simultaneously: \begin{enumerate*}[label=(\roman*)] \item the error in the model versus the number of iterations; \item the time spent per iteration. \end{enumerate*} Therefore, in this work we focus on studying the convergence rate with respect to the wall-clock time rather than the number of iterations. In fastest-$k$ SGD with fixed step size, the choice of the value of $k$ presents a trade-off between the convergence rate and the error floor. Namely, choosing a small value of $k$ will lead to fast convergence since the time spent per iteration would be short, however, this will also result in a low accuracy in the final model, i.e., higher error floor. Towards optimizing this trade-off, we study adaptive policies for fastest-$k$ SGD where the master starts with waiting for a small number of workers $k$ and then gradually increases $k$ to minimize the error as a function of time. Such an optimal adaptive policy would guarantee that the error is minimized at any instant of the wall-clock time. This would be particularly useful in applications where SGD is run with a deadline, since the learning algorithm would achieve the best accuracy within any time restriction.
 
\subsection{Related work} 
\subsubsection{Distributed SGD} The works that are closely related to our work are that of \cite{Gauri,CPMBJ16}. In \cite{CPMBJ16} the authors study fastest-$k$ SGD for a predetermined $k$. In \cite{Gauri}, the authors consider the same setting as \cite{CPMBJ16} and analyze the convergence rate of fastest-$k$ SGD with respect to the number of iterations. In addition to the convergence analysis with respect to the number of iterations, the authors in \cite{Gauri} separately analyze the time spent per iteration as a function of $k$. 

Several works proposed using redundancy when distributing the data to the workers. The master then uses coding theoretical tools to recover the gradient in the presence of a fixed number of stragglers, for example \cite{TLDK17,YA18,raviv2017gradient,lee2018speeding,ferdinand2018anytime,yu2018lagrange,kiani2018exploitation,chen2018draco,karakus2017straggler,halbawi2017improving,DCG16,KS18,fahim2017optimal,BPR17}. In \cite{charles2017approximate,CPE17,MRM18} the authors propose a mixed strategy in which the master distributes the data redundantly and uses coding techniques to obtain the whole gradient for a given number of stragglers. In addition, if more workers than accounted for are stragglers, the master can use the same coding techniques to compute an estimate of the gradient.

Note that the setting of the previously mentioned works, and the setting of interest for our work, focuses on the so-called synchronous SGD in which the workers are all synchronized at each iteration (i.e., have the same model). The literature also studies the asynchronous setting. In asynchronous distributed SGD, whenever a worker finishes its assigned computation, it sends the result to the master who directly updates $\bw$ and sends an updated $\bw$ to that worker who starts a new computation of the partial gradient while the other workers continue their previous computation, see for example \cite{Gauri,recht2011hogwild,liu2015asynchronous,shalev2013accelerated,reddi2015variance,pan2016cyclades}.

\subsubsection{Single-node SGD} Murata \cite{Murata} showed that irrespective of its convergence speed, the single-node SGD algorithm with fixed step size goes through a transient phase and a stationary phase. In the transient phase, $\bw_j$ approaches $\bw^\star$ exponentially fast in the number of iterations. Whereas, in the stationary phase, $\bw_j$ oscillates around $\bw^\star$. Note that  if a decreasing step size over the iterations is used, then $\bw_j$ converges to $\bw^\star$ rather than oscillating around it, however this leads to a long transient phase and hence a lower convergence rate. To detect the phase transition, \cite{chicago} uses a statistical test based on Pflug's method \cite{Pflug} for stochastic approximation. Detecting the phase transition serves many purposes, such as indicating when to stop the SGD algorithm or when to start implementing further tricks to reduce the distance between $\bw_j$ and $\bw^\star$. In this paper, we build on this line of work to derive the times at which the master should start waiting for more workers in fastest-$k$ SGD.

In another line of work on single-node SGD, the authors in~\cite{google} suggested increasing the batch size with the number of iterations as an alternative to decreasing the step size. The results in~\cite{google} show that increasing the batch size while keeping a constant step size, leads to near-identical model accuracy as decreasing the step size, but with fewer parameter updates, i.e., shorter training time.

\subsection{Our contributions}
We focus on straggler mitigation in synchronous distributed SGD with fixed step size. We consider a setting where the master distributes the data without redundancy. Under standard assumptions on the loss function, and assuming independent and identically distributed random response times for the workers, we give a theoretical bound on the error of fastest-$k$ SGD as a function of time rather than the number of iterations. We derive an adaptive policy which shows that this bound on the error can be optimized as a function of time by increasing the value of $k$ at specific times which we explicitly determine in terms of the system parameters. Furthermore, we develop an algorithm for adaptive fastest-$k$ SGD that is based on a statistical heuristic which works while being oblivious to the system parameters. We implement this algorithm and provide numerical simulations which show that the adaptive fastest-$k$ SGD can outperform both non-adaptive fastest-$k$ SGD and asynchronous SGD.

\section{Preliminaries}
In this paper we focus on fastest-$k$ SGD with fixed step size.  We consider a random straggling model where the time spent by worker $i$ to finish the computation of its partial gradient (i.e., response time) is a random variable $X_{i}$, for $i=1,\ldots,n$. We assume that $X_{i}, i=1,\ldots, n$, are \emph{iid} and independent across iterations. Therefore, the time per iteration for fastest-$k$ SGD is given by the $k^{th}$ order statistic of the random variables $X_1,\ldots,X_n$, denoted by~$X_{(k)}$. In the previously described setting, the following bound on the error of fastest-$k$ SGD as a function of the number of iterations was shown in~\cite{Gauri,Bottou}.
\begin{proposition}[Error vs. iterations of fastest-$k$ SGD \cite{Gauri, Bottou}]
\label{prop1}
Under certain assumptions (stated in~\cite{Gauri,Bottou}), the error of fastest-$k$ SGD after $j$ iterations with fixed step size satisfies
\begin{equation*}
\mathbb{E}\left[F(\mathbf{w}_j)-F^{\star} \right]\leq \frac{\eta L \sigma^2}{2cks} +(1-\eta c)^{j}  \bigg(F(\mathbf{w}_0)-F^{\star}-\frac{\eta L \sigma^2}{2cks} \bigg) \label{eeq10},
\end{equation*}
where $L$ and $c$ are the Lipschitz and the strong convexity parameters of the loss function respectively, $F^{\star}$ is the optimal value of the loss function, and $\sigma^2$ is the variance bound on the gradient estimate.
\end{proposition}

\section{Theoretical Analysis}
In this section, we present our theoretical results. The proofs of these results are available in~\cite{Ex}. In Lemma~\ref{lem1}, by applying techniques from renewal theory, we give a bound on the error of fastest-$k$ SGD as a function of the wall-clock time $t$ rather than the number of iterations. The bound holds with high probability for large $t$ and is based on Proposition~\ref{prop1}.
\begin{lemma} [Error vs. wall-clock time of fastest-$k$ SGD] 
\label{lem1}
Under the same assumptions as Proposition~\ref{prop1}, the error of fastest-$k$ SGD after wall-clock time $t$ with fixed step size satisfies
\begin{align}
\label{eeq9}
&\mathbb{E}\left[F(\mathbf{w}_t)-F^{\star} \right | J(t)]\leq  \frac{\eta L \sigma^2}{2cks}  \nonumber \\ 
 & \hspace{1cm} + (1-\eta c)^{\frac{t}{\mu_k}(1-\epsilon)}  \left(F(\mathbf{w}_0)-F^{\star}-\frac{\eta L \sigma^2}{2cks} \right),
\end{align}
with high probability \big($Pr \geq 1 - \frac{\sigma_k^2}{\epsilon^2}(\frac{2}{t\mu_k}+\frac{1}{t^2}) $\big) for large $t$, where $0<\epsilon \ll 1$ is a constant error term, $J(t)$ is the number of iterations completed in time $t$, and $\mu_k$ is the average of the $k^{th}$ order statistic $X_{(k)}$.
\end{lemma}

Notice that the first term in~\eqref{eeq9} is constant (independent of $t$), whereas the second term decreases exponentially in~$t$ ($\eta c<1$ from~\cite{Bottou}). In fact, it is well-known that SGD with constant step size goes first through a transient phase where the error decreases exponentially fast, and then enters a stationary phase where the error oscillates around a constant term~\cite{Murata}. From~\eqref{eeq9}, it is easy to see that the rate of the exponential decrease in the transient phase is governed by the value of $1/\mu_k$. $\mu_k$ is an increasing function of $k$, thus the exponential decrease is fastest for $k=1$ and slowest for $k=n$. Whereas the stationary phase error which is upper bounded by $\eta L \sigma^2/2cks$, is highest for $k=1$ and lowest for $k=n$. This creates a trade-off between the rate of decrease of the error in the transient phase, and the error floor achieved in the stationary phase. Ultimately, we would like to first have a fast decrease through the transient phase, and then have a low error in the stationary phase. To this end, we look for an adaptive policy for varying $k$ that starts with $k=1$ and then switches to higher values of $k$ at specific times in order to optimize the error-runtime trade-off. Such an adaptive policy guarantees that the error is minimized at every instant of the wall-clock time $t$.

Since the bound in~\eqref{eeq9} holds with high probability for large~$t$, we explicitly derive the switching times that optimize this bound. Note that for the sake of simplicity, we drop  the constant error term $\epsilon$ in our analysis.

\begin{theorem}[Bound-optimal Policy]
\label{thm2}
The bound-optimal times $t_k, k=1, \ldots, n-1$, at which the master should switch from waiting for the fastest $k$ workers to waiting for the fastest $k+1$ workers are given by
\begin{align*}
& t_k  = t_{k-1}+\frac{\mu_k}{-\ln (1-\eta c)} \times  \bigg[ \ln \left(\mu_{k+1}-\mu_k \right)   -  \ln \left(\eta L \sigma^2\mu_k \right) \nonumber \\
&+\ln \left(2ck(k+1)s(F(\mathbf{w}_{t_{k-1}})-F^{\star})-\eta L (k+1) \sigma^2) \right)\bigg] \label{eqq11},
\end{align*} 
where $t_0=0$.
\end{theorem}

\begin{example}[Theoretical analysis on adaptive fastest-$k$ SGD with {\em iid} exponential response times] 
\label{ex2}
 Suppose $X_i\sim \exp(\mu),~i=1,\ldots,n$. The average time spent per iteration is $\mu_k=H_n-H_{n-k}$, where $H_n$ is the harmonic number. Let $n=5, \mu=5, \eta=0.001, \sigma^2=10, F(\mathbf{w}_0)-F^{\star}=100, L=2, c=1, s=10$. We evaluate the bound in Lemma~\ref{lem1} for multiple fixed values of $k$ (non-adaptive) and compare it to adaptive fastest-$k$ SGD if we apply the switching times in Theorem~\ref{thm2}. The results are shown in Fig.~\ref{fig2}. 

\begin{figure}[h]
\vspace{-0.2cm}
\centering
\includegraphics[width=0.33\textwidth]{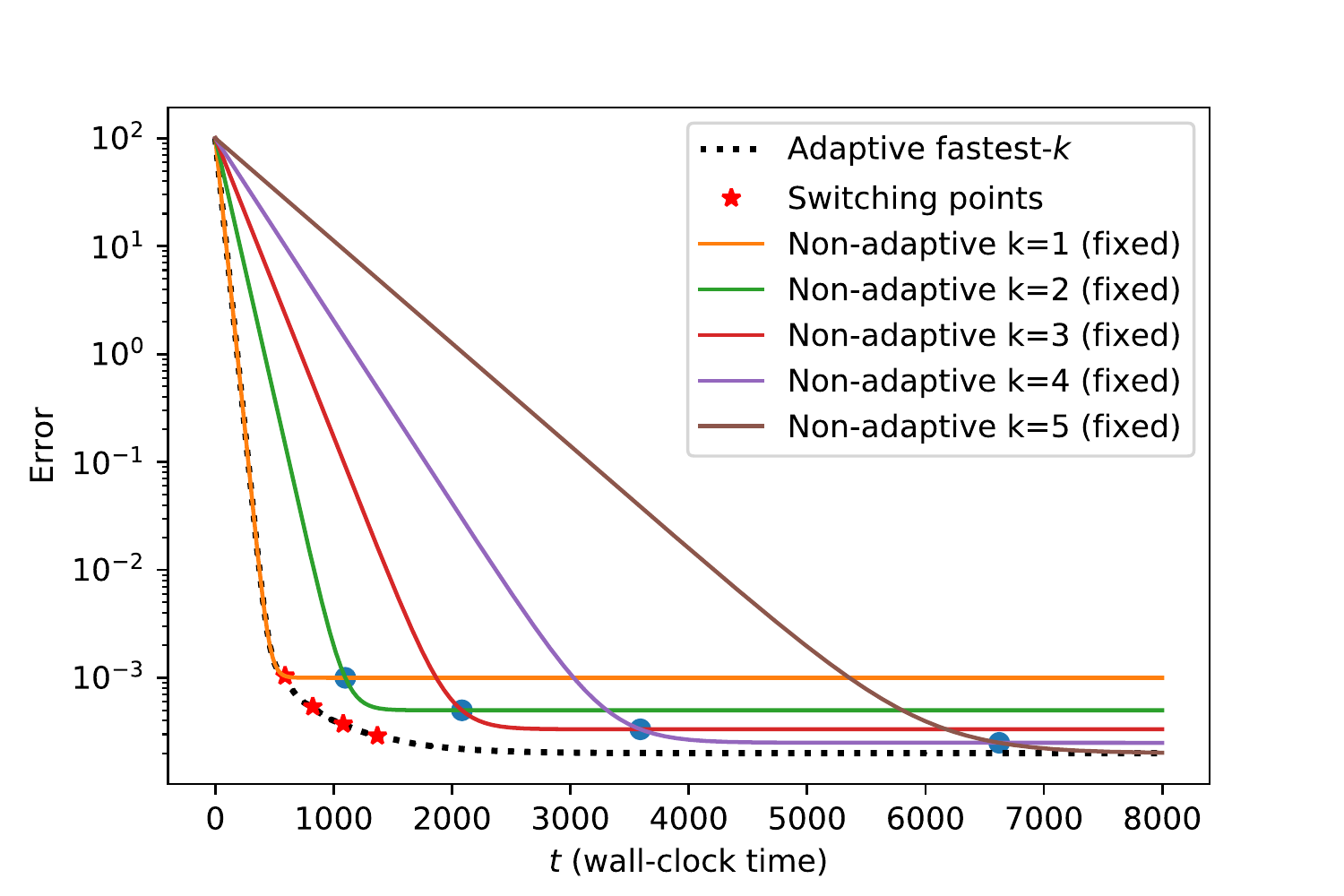}
\caption{The upper bound on the error given by~\eqref{eeq9} as a function of time, evaluated for $k=1,2,3,4,5$.}
\label{fig2}
\vspace{-0.2cm}
\end{figure}

Notice from Fig.~\ref{fig2} that for the time interval $[0,t_1)$, the adaptive policy assigns $k=1$ since it gives the fastest error decrease in the beginning. Then, as the error approaches the stationary phase, the policy increases $k$ to $k=2$. This allows the error to decrease below the error floor for $k=1$. The procedure continues until $k$ attains its maximum value of $k=n=5$. The results demonstrate the adaptive version enables achieving lower error values in less time.
\end{example}

This analysis shows the potential of adaptive strategies in optimizing the error-runtime trade-off. It also suggests that the value of $k$ should be gradually increased throughout the runtime of fastest-$k$ SGD in order to optimize this trade-off. Although this analysis provides useful insights about how to adapt $k$ over time, it may not be effective in practice for the following two reasons: \begin{enumerate*}[label={(\roman*)}] \item the policy optimizes an upper bound on the error (Lemma~\ref{lem1}) which is probabilistic and may be loose; \item the policy requires the knowledge of several system parameters including the optimal value of the loss function $F^{\star}$ which is typically unknown.\end{enumerate*} Nevertheless, we use the insights provided by the theoretical analysis to design a practical algorithm for adaptive fastest-$k$ SGD. This algorithm is based on a statistical heuristic and is oblivious to the system parameters as we explain in Section~\ref{sec5}. 

\section{Adaptive fastest-$k$ SGD Algorithm}
\label{sec5}
In this section we present an algorithm for adaptive fastest-$k$ SGD that is realizable in practice. As previously mentioned, SGD with fixed step size goes first through a transient phase where the error decreases exponentially fast, and then enters a stationary phase where the error oscillates around a constant term. Initially, the exponential decrease is fastest for $k=1$. Then, as the stationary phase approaches, the error decrease becomes slower and slower until a point where the error starts oscillating around a constant term and does not decrease any further. At this point, increasing $k$ allows the error to decrease further because the master would receive more partial gradients and hence would obtain a better estimate of the full gradient. The goal of the adaptive policy is to detect this phase transition in order to increase $k$ and keep the error decreasing.

The adaptive policy we present in this section detects this phase transition by employing a statistical test based on a modified version of Pflug's procedure for stochastic approximation~\cite{Pflug}. The main component of our policy is to monitor the signs of the products of consecutive gradients computed by the master based on~\eqref{eeq6}. The underlying idea is that in the transient phase, due to the exponential decrease of the error, the gradients are likely to point in the same direction, hence their inner product is positive. Our policy consists of utilizing a counter that counts the difference between the number of times the product of consecutive gradients is negative (i.e., ${\ghat_j}^T \ghat_{j-1} < 0$) and the number of times this product is positive, throughout the iterations of the algorithm. 

In the beginning of the algorithm, we expect the value of the counter to be negative and decrease because of the exponential decrease in the error. Then, as the error starts moving towards the stationary phase, negative gradient products will start accumulating until the value of the counter becomes larger than a certain positive threshold. At this point, we declare a phase transition and increase $k$. The complete algorithm is given in Algorithm~\ref{algo_disjdecomp}.

\begin{algorithm}[h]
\scriptsize
\SetKwData{Left}{left}
\SetKwData{This}{this}
\SetKwData{Money}{money}
\SetKwData{MaxIter}{maxIter}
\SetKwData{Step}{step}
\SetKwData{CountI}{countIter}
\SetKwData{CountN}{countNegative}
\SetKwData{Thresh}{thresh}
\SetKwData{Burnin}{burnin}
\SetKwData{Up}{up}
\SetKwFunction{Budget}{budget}
\SetKwFunction{Union}{Union}
\SetKwFunction{FindCompress}{FindCompress}
\SetKwInOut{Input}{input}
\SetKwInOut{Output}{output}
\Input{starting point $\mathbf{w}_0$, data $\{X,\mathbf{y}\}$, number of workers $n$, step size $\eta$, maximum number of iterations~$J$, adaptation parameters \Step , \Thresh, \Burnin }
\Output{weight vector $\mathbf{w}_J$}
$j\gets 1$ \\
$k \gets 1$ \\
$\CountN \gets 0$ \\
$\CountI \gets 1$ \\
Distribute $X$ to the $n$ workers \\
\While{$j\leq J$}{
Send $\mathbf{w}_{j-1}$ to all workers \\
Collect the responses of the fastest $k$ workers \\
$\mathbf{w}_{j}\gets \mathbf{w}_{j-1}-\eta \ghat_{j-1}$ \\
\eIf{${\ghat_j}^T \ghat_{j-1}  < 0$ } { 
$\CountN \gets \CountN + 1$} { $\CountN \gets \CountN - 1$}

\If{$\CountN>\Thresh$ {\bf and} $\CountI>\Burnin$ {\bf and} $k\leq n-\Step$} {
$k\gets k+\Step$ \\
$\CountN\gets 0$ \\
$\CountI\gets 0$ 
}
$\CountI \gets \CountI + 1$ \\
$j\gets j+1$ 
}
\Return $\mathbf{w}_J$ 
\label{alg1}
\caption{Adaptive fastest-$k$ SGD}\label{algo_disjdecomp}
\end{algorithm}

\section{Simulations}
\label{sec6}
\subsection{Experimental setup}
We simulated the performance of the adaptive fastest-$k$ SGD (Algorithm~\ref{algo_disjdecomp}) described earlier for $n$ workers on synthetic data $X$. We generated $X$ as follows: \begin{enumerate*}[label={(\roman*)}] \item we pick each row vector $\mathbf{x}_{\ell}$, $\ell=1,\dots,m,$ independently and uniformly at random from $\{1,2,\dots,10\}^{d}$;  \item we pick a random vector $\bar{\mathbf{w}}$ with entries being integers chosen uniformly at random from $\{1,\dots,100\}$; and \item we generate $\mathbf{y}_{\ell} \sim \mathcal{N}(\ip{\mathbf{x}_{\ell}}{\bar{\mathbf{w}}},1)$ for all $\ell=1,\dots,m$.\end{enumerate*} We run linear regression using the $\ell_2$ loss function. At each iteration, we generate $n$ independent exponential random variables with rate $\mu = 1$. 

\subsection{ Adaptive fastest-$k$ SGD vs Non-adaptive fastest-$k$ SGD}
\begin{figure}[h!]
\centering
\includegraphics[width=0.36\textwidth]{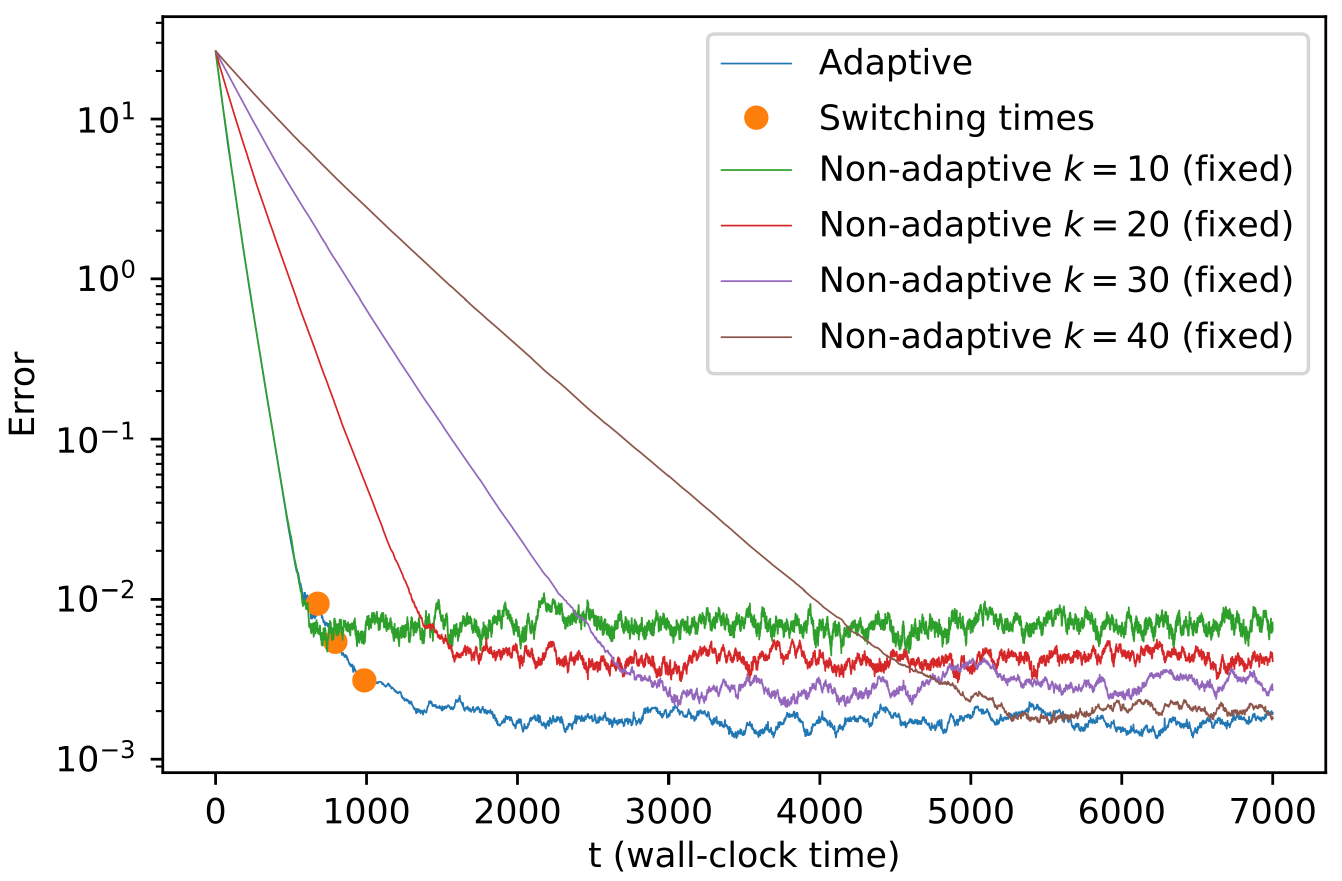}
\caption{\footnotesize Error as a function of the wall-clock time for non-adaptive fastest-$k$ SGD with fixed $k=10,20,30,40$; and adaptive fastest-$k$ SGD (Algorithm~1). The experimental setup is the following: $d=100$, $m=2000$, $n=50$, $\eta=0.0005$. The adaptation parameters chosen here are $step=10$, $thresh=10$, and $burnin=0.1\times $(number of data points) $=200$. We start the adaptive fastest-$k$ SGD with $k=10$ and increase $k$ by $10$ until reaching $k=40$.}
\label{fig3}
\vspace{-0.4cm}
\end{figure}

Figure~\ref{fig3} compares the performance of the adaptive fastest-$k$ SGD to non-adaptive for $n=50$ workers. In the adaptive version we start with $k=10$ and then increase $k$ by $10$ until reaching $k=40$, where the switching times are given by Algorithm~\ref{algo_disjdecomp}. Whereas for the non-adaptive version, $k$ is fixed throughout the runtime of the algorithm. The comparison shows that the adaptive version is able to achieve a better error-runtime trade-off than the non-adaptive one. Namely, notice that the adaptive $k$-sync reaches its lowest error at approximately $t=2000$, whereas the non-adaptive version reaches the same error only for $k=40$ at approximately $t=6000$. These results confirm our intuition and previous theoretical results.

\subsection{Comparison to Asynchronous SGD}
\begin{figure}[hhh!]
\centering
\includegraphics[width=0.36\textwidth]{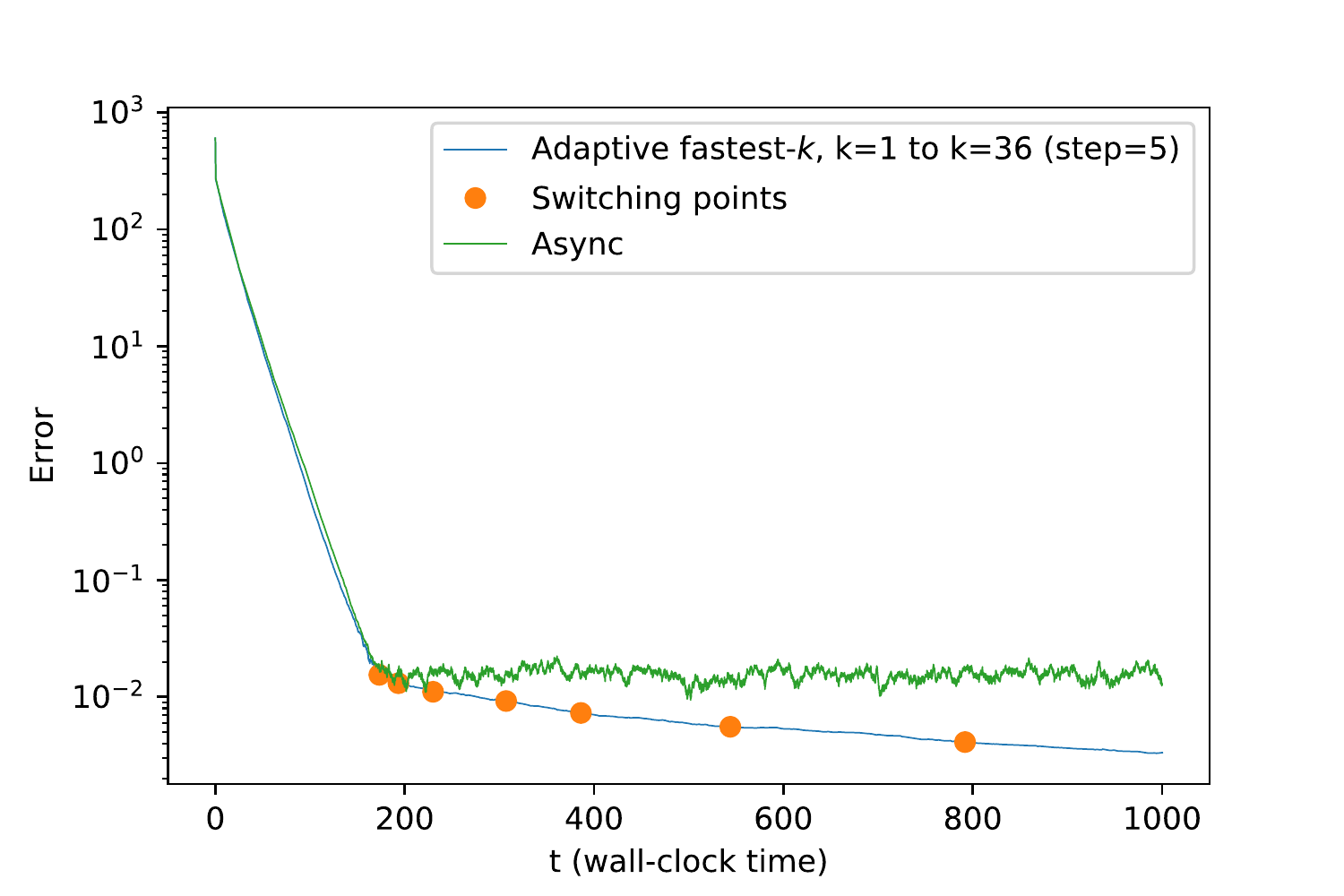}
\caption{\footnotesize Error as a function of time for adaptive fastest-$k$ SGD (Algorithm~1) and asynchronous SGD. The experimental setup is the following: $d=100$, $m=2000$, $n=50$, $\eta=0.0002$. The adaptation parameters chosen here are $step=5$, $thresh=10$, and $burnin=0.1\times $(number of data \mbox{points) $=200$}. We start the adaptive fastest-$k$ SGD with $k=1$ and increase $k$ by $5$ until reaching $k=36$.
}
\label{fig4}
\vspace{-0.4cm}
\end{figure}

Figure~\ref{fig4} compares the adaptive fastest-$k$ SGD to the fully asynchronous version of distributed SGD~\cite{Gauri}. Similar conclusions can be drawn as in the case of Figure~\ref{fig3}.


\bibliography{Refs}
\bibliographystyle{ieeetr}

\end{document}